%% file: paper.tex
\documentclass[twoside]{article}
\usepackage[accepted]{aistats2018}
\usepackage{natbib}
\setcitestyle{round}

\usepackage{graphicx} 
\usepackage{subfigure} 
\usepackage{array}
\usepackage{tabulary}
\newcolumntype{K}[1]{>{\centering\arraybackslash}p{#1}}
\usepackage{hhline}

\usepackage{algorithm}
\usepackage{algorithmic}
\usepackage{multirow}
\usepackage{times}

\usepackage{bm}
\usepackage{amsmath}
\usepackage{amssymb}
\usepackage{amsthm}
\usepackage[textsize=tiny]{todonotes}
\usepackage{stfloats}
\usepackage{enumitem}
\usepackage[flushleft]{threeparttable}
\usepackage{subfig}
\usepackage{wrapfig}

\newtheorem{definition}{Definition}
\newtheorem{theorem}{Theorem}

\def\argmax{\operatornamewithlimits{argmax}}

\begin{document}

%

%

\twocolumn[

\aistatstitle{Robust Covariate Shift Prediction with General Losses and Feature Views}

\aistatsauthor{ Anqi Liu \And Brian D. Ziebart }

\aistatsaddress{ University of Illinois at Chicago\And University of Illinois at Chicago} ]

\begin{abstract}
Covariate shift 
relaxes the widely-employed 
\emph{independent and identically distributed} (IID) assumption by allowing
different training and testing input distributions.
Unfortunately, common methods for addressing covariate shift by trying to 
remove the bias between training and testing distributions using
importance weighting often provide poor performance guarantees in theory
and unreliable predictions with high variance in practice. 
Recently developed methods that construct a predictor
that is inherently robust to the difficulties of learning under covariate shift
are restricted to minimizing logloss and can be too conservative when faced with high-dimensional learning tasks.
We address these limitations in two ways: 
by robustly minimizing various loss functions, including non-convex ones, under the testing distribution; 
and by separately shaping the influence of covariate shift according to different 
feature-based views of the relationship between input variables and example
labels. 
These generalizations make robust covariate shift prediction applicable to more task scenarios.
We demonstrate the benefits on classification under
covariate shift tasks.
\end{abstract}

\section{Introduction}

The \emph{independent and identically distributed} (IID) 
assumption employed widely
across machine learning methods requires the testing data distribution to be the same as the training data distribution. This is quite restrictive in the sense that shift can occur between the training distribution and testing
distribution in many settings, which makes models built on the IID assumption inappropriate.
Specifically, the predictor minimizing (regularized) loss defined on the training 
samples provides no performance guarantees when applied to the
testing distribution \citep{zadrozny2004learning, fan2005improved}.
Though nothing can be learned when the shift between training and testing
data is arbitrary, certain assumptions about how training and testing distributions differ allow reasonable adaptive learning methods to be derived \citep{blitzer2008learning}. One of the common assumptions is that the bias just comes from the input variables.  In this setting, also known as 
{\bf covariate shift}, only the distribution of inputs, 
$P_{\text{train}}({\bf x})$ and $P_{\text{test}}({\bf x})$, differ,
while the conditional label distribution, $P(y|{\bf x})$, is the same under
both the training and the testing distributions. This 
assumption is much weaker than the IID assumption and covers a broad range of real application scenarios.

The most prevalent methods for addressing covariate shift attempt to debias
the training data by reweighting it using a density ratio,
$P_{\text{test}}({\bf x})/ P_{\text{train}}({\bf x})$.  
This approach tends to work well when the training and the testing
distributions are fairly similar and large amounts of training samples are
available. It also enjoys asymptotical consistency guarantee provided an 
infinite amount of training data.
However, when these conditions are violated, i.e., there is only a limited 
amount of training data and/or significant differences
between the training and testing distributions, 
some of the density ratios
for training examples can be extremely large.
This leads to high-variance
estimates that extrapolate heavily from scant amounts of training data 
and a lack of generalization guarantees for the resulting
predictor \citep{cortes2010learning,cortes2008sample}.

Recently developed robust covariate shift methods take a worst-case approach,
constructing a predictor that (approximately) matches training data statistics,
but is otherwise the most uncertain on the testing distribution
\citep{liu2014robust,chen2016robust}. These methods were built by minimizing the worst case expected target logloss and obtain a 
parametric form of the predicted output labels' probability distributions. 
Unfortunately, log loss may not be of interest for many applications and robust accuracy maximization is instead desired, for example.
In addition, these previous methods can be \emph{too} robust, providing overly
conservative predictions that are nearly uniform on portions of the testing input space that lacks training support, particularly when 
the dimensionality of the input space is large---a situations that is
problematic for importance weighted loss minimization as well.
In this paper, we introduce two generalization 
of robust covariate shift classification.
First, we robustly minimize other loss functions, like the 0-1 loss, under covariate shift. 
Second, we enable a better balance between making generalization assumptions about data and providing
robustness to uncertainty by introducing feature view-based generalization
assumptions to the robust covariate shift approach. We demonstrate the effectiveness of our method using both synthetic examples and real datasets.

\section{Background}

\subsection{Covariate Shift}

In many learning settings, the distribution available for training a predictor
differs from the distribution of data on which it will be employed and
evaluated.
Under covariate shift, the training distribution and testing distribution
share the same conditional label distribution, $P(y|{\bf x})$, but have
differing distributions over inputs:
\begin{align}
& P_{\text{train}}({\bf x}, y) =  P_{\text{train}}({\bf x}) P(y|\bf{x}) \notag\\
&P_{\text{test}}({\bf x}, y) =  P_{\text{test}}({\bf x}) P(y|\bf{x}).
\end{align}
Unlike the more common \emph{independent and identically distributed} (IID)
learning setting \citep{mohri2012foundations}, 
which further imposes that 
$P_{\text{train}}({\bf x}) = P_{\text{test}}({\bf x})$,
constructing a predictor with limited complexity that performs well on the 
training data does not guarantee good performance on the testing distribution
(with high probability).
Also known as sample selection bias \citep{heckman1977sample}, covariate shift has ties to more general domain adaptation \citep{jiang2008literature} and transfer learning settings \citep{pan2010survey}, where the assumption on how training and testing data distributions differ is not fully specified. 

\subsection{Debiasing Via Importance Weighting}

The most 
prevalent approach for addressing covariate shift 
attempts to 
remove the bias between the training and testing distributions
\citep{shimodaira2000improving, huang2006correcting, sugiyama2008direct}.
Under this perspective, minimizing the 
importance-weighted loss of $(n)$ training examples, 
\begin{align}
& \lim_{n \rightarrow \infty} \min_{\hat{f}}\mathbb{E}_{({\bf X},Y)\sim
\tilde{P}_{\text{train}}^{(n)}}\left[
\frac{P_{\text{test}}({\bf X})}{P_{\text{train}}({\bf X})}
\text{loss}(\hat{f}({\bf X}),Y) \right]\notag\\ 
&=  \min_{\hat{f}} \mathbb{E}_{({\bf X},Y)\sim
P_{\text{test}}}\left[
\text{loss}(\hat{f}({\bf X}),Y) \right], \label{eq:reweight}
\end{align}
where $\hat{f}$ is estimated predictor and $\tilde{p}$ is the empirical distribution of data, asymptotically minimizes the testing distribution loss, so long as 
$P_{\text{test}}({\bf x}) > 0 \implies P_{\text{train}}({\bf x}) > 0$.  

Despite this asymptotic guarantee, predictive performance can be poor when
training from finite amounts of samples in both theory and practice.
Conceptually, the density ratios of a small number of training examples
can become disproportionately large, making the resulting predictor overly
sensitive to a small number of training data points---or even one
single datapoint. This will lead to predictive results with high variance. 
Indeed, finite generalization bounds for importance-weighted methods
require finite second moments: $\mathbb{E}_{P_{\text{train}}(x)}[(P_{\text{test}}({\bf X})/P_{\text{train}}({\bf X}))^2] < \infty$ 
\citep{cortes2010learning}, which is often not satisfied in practice.

In order to overcome these difficulties, there is a significant 
literature studying how to reasonably estimate the weight $\frac{P_{\text{test}}({\bf x})}{P_{\text{train}}({\bf x})}$ from training and testing sample data \citep{gretton2009covariate}. 
For example, methods based on minimizing certain types of divergences or loss functions on density (ratios) \citep{sugiyama2008direct, kanamori2009efficient}
have been investigated. Other methods \citep{cortes2014domain} make implicit assumptions on the feature space that there exist weights $w(x)$ that makes the ``distance'' between training and testing features small enough so that reweighted training features can be used to learn classifiers on testing distribution. Other work focuses on utilizing two stages of regularization to reduce the variance of the resulting predictions \citep{reddi2014doubly}.

 \subsection{Robust Bias-Aware Prediction}
Robust covariate shift classification \citep{liu2014robust} is motivated from minimax robust estimation \citep{topsoe1979information, grunwald2004game}.
The expected testing loss minimization is formulated as a two player game, where the estimator player seeks to minimize the loss function, while an adversarial player tries to maximize it under constraints based on training samples. 
\begin{align}
\min_{\hat{P}} \max_{\check{P} \in \tilde{\Xi}_{\text{train}}}
\mathbb{E}_{{\bf X} \sim P_{\text{test}},
\check{Y}|{\bf X} \sim \check{P}}
\left[-\log \hat{P}(\check{Y}|{\bf X})\right],\label{eq:value}
\end{align}
The adversary must choose a distribution $\check{P}$ that is similar to 
certain measured properties (features), e.g., 
$\mathbb{E}_{{\bf X} \sim \tilde{P}, \check{Y}|{\bf X} \sim \check{P}}
\left[\phi({\bf X},\check{Y})\right] = 
\mathbb{E}_{({\bf X},Y) \sim \tilde{P}}\left[\phi({\bf X},Y) \right]$, 
of the training data.
These are denoted by the convex set $\tilde{\Xi}_{\text{train}}$, with $\phi$ as the feature function.
The advantage of this formulation resides in its robustness to the worst possible case of covariate shift and avoidance of huge losses caused by over optimistic extrapolations.  Moreover, the expected testing loss under this formulation is upper bounded by the testing entropy \citep{liu2015shift}. 

Recent advances in adversarial loss minimization in the IID setting extend beyond the logloss to non-smooth loss functions, such as the 0-1 loss \citep{NIPS2016_6088, NIPS2016_6247} and ordinal regression loss \citep{ordinal}. 
In this paper, we establish the general form of the method that minimizes different loss functions under covariate shift. 


\section{Generalized Adversarial Covariate Shift}

We now provide a general form of the adversarial game for covariate shift that incorporates a wider set of loss functions and transfer assumptions between training and testing distributions. We then introduce variants and special cases that 
are applicable for different scenarios.

\subsection{General Loss and Feature Shift Formulation}


Our goal is to construct a predictor that is robust to the worst case testing distribution implied by available training data.
To allow flexibility in what the training data can imply, 
we assume there exists a generalization distribution, $P_{\text{gen}_v}({\bf x})$, where features of training data are assumed to generalize.
We consider a set $\mathcal{V}$ 
of view-based feature vectors, 
$\phi_v({\bf x}_v, y)$, for $v \in \{1, \hdots, |\mathcal{V}|\}$, 
defined over a subset of the input variables ${\bf x}_v$. 
We then apply the robust method for covariate shift to the generalization 
distribution instead of the original training distribution $P_{\text{train}}({\bf x})$,
as shown in the following definition:
\begin{definition} \label{def:generalgame}
The {\bf generalized robust covariate shift classifier} results from
the adversarial loss optimization game:
\begin{align}
& \min_{\hat{P}} \max_{\check{P}}
\mathbb{E}_{{\bf X} \sim P_{\text{test}}}
\left[\text{Loss}(\check{P}_{\bf X}, \hat{P}_{\bf X})\right] \text{ such that: } \label{eq:generalgame}\\
& \forall v \in \{1, \hdots, |\mathcal{V}|\}, 
\;\;
\mathbb{E}_{{\bf X} \sim \tilde{P}_{\text{gen}_v}, \check{Y}|{\bf X} \sim
\check{P} }
\left[
\phi_v({\bf X}_v,\check{Y}) \right] 
=  \notag \\ & \qquad\qquad\qquad\qquad
\mathbb{E}_{({\bf X},Y) \sim \tilde{P}_{\text{train}}}
\left[\frac{P_{\text{gen}_v}(X)}{P_{\text{train}}(X)}
\phi_v({\bf X}_v,Y) \right], \notag
\end{align}
with a loss function that we want to minimize and a distribution, $P_{\text{gen}_v}(X)$, to which training data feature statistics are assumed to generalize.
\end{definition}

Note that the statistics in the constraints 
are reweighted training sample statistics. 
This provides us with flexibility to impose different assumptions---in terms of $P_{\text{gen}_v}({\bf x})$ densities---for how training data should generalize.

Strong Lagrangian duality holds when $\text{Loss}(\cdot,\cdot)$ is a concave-convex function of $\check{P}$ and $\hat{P}$.  
This enables us to re-write the game in terms of 
Lagrangian multipliers $\theta$: 
\begin{align}
& \min_{\theta}\min_{\hat{P}} \max_{\check{P}}
\mathbb{E}_{{\bf X} \sim P_{\text{test}}}
\Bigg[\text{Loss}(\check{P}_{\bf X}, \hat{P}_{\bf X}) \label{eq:obj} \\
& \quad
+ \sum_v\frac{P_{\text{gen}_v}(X)}{P_{\text{test}}(X)}\theta_v\phi_v(X_v, \check{Y})\Bigg] - \sum_v\theta_v\tilde{\phi_v} + \epsilon ||\theta||_2, \notag
\end{align}
where $\tilde{\phi}_v \triangleq \mathbb{E}_{({\bf X},Y) \sim \tilde{P}_{\text{train}}}
\left[\frac{P_{\text{gen}_v}({\bf X})}{P_{\text{train}}({\bf X})}
\phi_v({\bf X}_v,Y) \right]$ is the feature function evaluated on empirical training data, and we allow $\epsilon$ slack for matching the primal constraints, leading to regularization in the dual. 
The optimization of this objective function is then composed of two steps: first, solve the
 inner minimax game with respect to $\hat{P}$ and $\check{P}$; second, optimize for $\theta$ in the outer minimization to satisfy imposed constraints. We focus our attention on $0-1$ loss and log loss, but many other loss functions can also be incorporated.

\subsection{Classification Losses} Letting $\text{Loss}(\check{P}_{\bf X}, \hat{P}_{\bf X}) = \hat{P}^{T}C\check{P}$---a bilinear and therefore concave-convex function of $\check{P}$ and $\hat{P}$---allows many classification losses to be represented in the cost matrix C.
We can reformulate the inner minimax game as $\min_{\hat{P}} \max_{\check{P}}
\mathbb{E}_{{\bf X}}[\hat{P}^{T}C'\check{P}]$, where $C' = C + \sum_v \frac{P_{\text{gen}_v}(X)}{P_{\text{test}}(X)}\theta_v\phi_v(X_v, \check{Y})$. The inner minimax game, which is a two player zero sum game, can be solved by linear programming. Another way to  find the equilibrium of the inner minimax game for the special case of 0-1 loss is by seeking an analytical form of the game value as in \cite{NIPS2016_6088}, which brings more computational efficiency. For the outer minimization, we take the subgradient with respect to $\theta$, which we approximate by reweighting training samples to the generalization distribution,
\begin{align}
\mathbb{E}_{{\bf X} \sim P_{\text{gen}_v}, \check{Y}|{\bf X}\sim \check{P}}\left[\phi_v({\bf X}_v, \check{Y})\right] - \tilde{\phi}_v + 2\epsilon\theta,
\end{align}
and perform subgradient descent. 

We are also able to bound the approximation error in both subgradient and objective, given the probability densities are accurate. We refer to more details in Section \ref{sec:bound} and just show two illustrative examples in Figure \ref{figs:zero_one}, where training distribution (solid line) and testing distribution(dashed line) is overlapping in different ways. The prediction color map shows a similar uncertain prediction with logloss-based classifier where there is not enough training data support, like the top right corner in the first figure. Moreover, the 0-1 loss 
provides more certain prediction in the overlapped region while logloss-based classifier's prediction changes gradually in certainty from the most supported region to the least.
{
\begin{figure}[htp]
\begin{tabular}{cc}
\includegraphics[height=3.5cm, trim  = 4cm 0cm 7cm 0.8cm, clip]{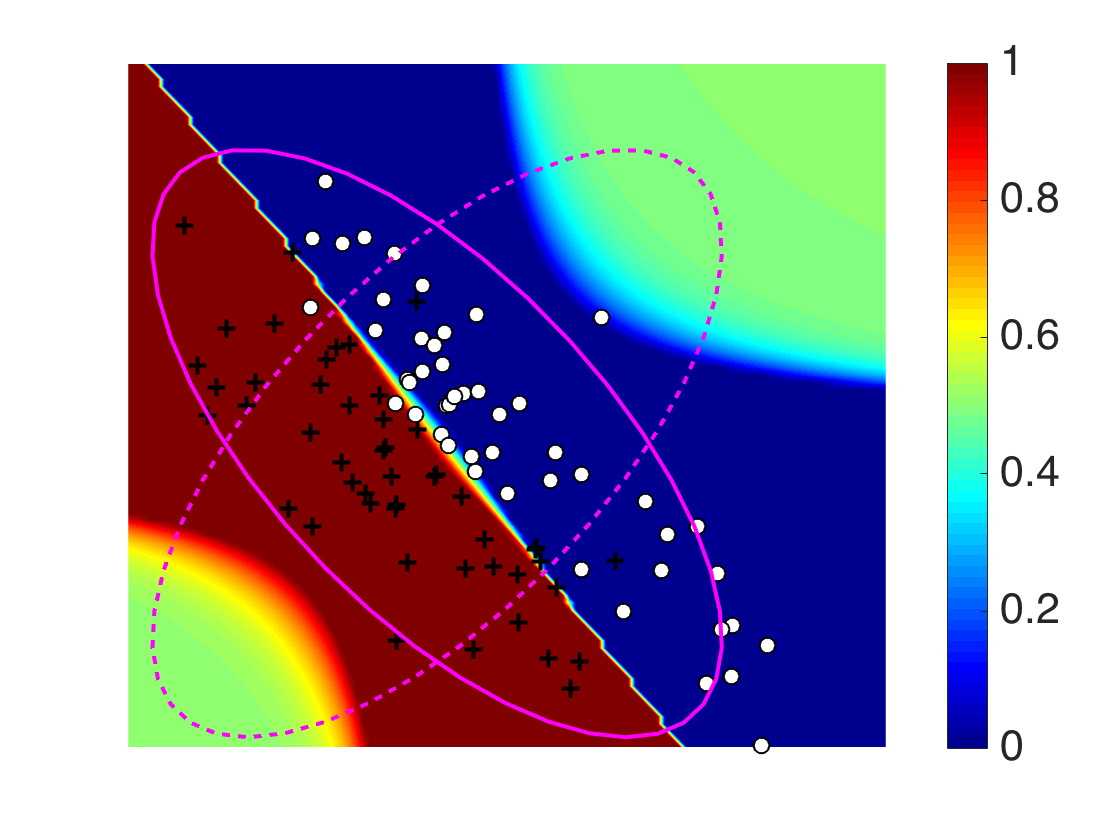} &
 \includegraphics[height=3.5cm, trim  = 4cm 0cm 2cm 0.8cm, clip]{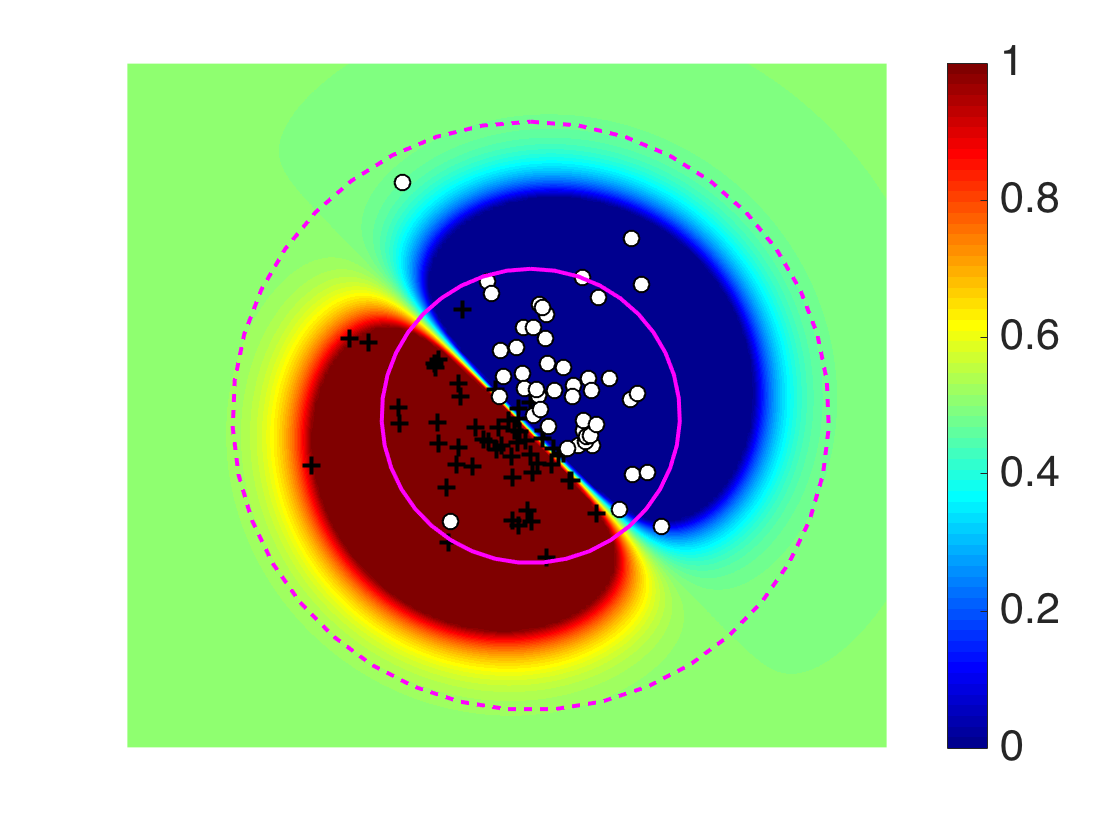}  \\
\end{tabular}
\caption{Prediction colormap with robust classifier using 0-1 loss when $P_{\text{gen}_v}({\bf x}) = P_{\text{train}}({\bf x})$. The colormap shows the $P($`+ '$ |{\bf x})$. Training data with 5\% noise is also shown.}
\label{figs:zero_one}
\end{figure}
}

\subsection{The Log Loss Setting}
When $\text{Loss}(\check{P}_{\bf X}, \hat{P}_{\bf X}) = \mathbb{E}_{\check{Y}|{\bf X} \sim \check{P}} [-\log \hat{P}(\check{Y}|{\bf X})]$, this is the logloss case which generalizes to multiple feature views. In the logloss case, the inner minimax game has the solution $\hat{P} = \check{P}$, so we follow the same procedure in \citep{liu2014robust} and establish the parametric form of the conditional
label distribution
in Theorem \ref{th:general}.

\begin{theorem} \label{th:general}The generalized robust covariate shift classifier based on logloss has the
following parametric form:
\begin{align}
\hat{P}_\theta(y|{\bf x}) \propto
e^{ \sum_v \frac{P_{\text{gen}_v}({\bf X})}{P_{\text{test}}({\bf x})} 
\theta_v \cdot \phi_v({\bf x}_v, y)}
.\label{eq:robust}
\end{align}
\end{theorem}

We need to specify $P_{\text{gen}_v}(x)$ carefully in practice (\S\ref{sec:role}), but first consider two special cases under logloss.

\paragraph{Reduction to Robust Bias-Aware Prediction:} 
When $P_{\text{gen}_v}({\bf x}) = P_{\text{train}}({\bf x})$, this gives us a model that is representationally equivalent to the robust bias aware prediction method proposed in \cite{liu2014robust}. The solution has a parametric form with the
density ratio appearing as:
$P(y|{\bf x}) \propto e^{
\frac{P_{\text{train}}({\bf x})}{P_{\text{test}}({\bf x})}
\sum_v \theta_v \cdot\phi_v({\bf x}_v,y)}$
and moderates the uncertainty of the predictor to be larger for inputs that
are relatively less likely in the training data. Thus, the density ratio 
$\frac{P_{\text{train}}({\bf x})}{P_{\text{test}}({\bf x})}$ in this parametric form controls the uncertainty of 
predictions.
Unfortunately, when the input space is high-dimensional this robustness
guarantee can be \emph{too} conservative and restrictive, leading to predictions that
are not informative (i.e., uniform distributions). 
This method is the most conservative model this framework generates, which limits the adversarial player to match training sample statistics strictly.

\paragraph{Reduction to Importance Weighting:}
If all the features are assumed to generalized fully to the testing distribution, i.e.,
$P_{\text{gen}_v}({\bf x}) = P_{\text{test}}({\bf x})$, 
the generalized robust covariate shift classifier is equivalent
to the importance weighting method (Eq. \ref{eq:reweight}). It produces the most aggressive model when the $\check{P}$ could match reweighted features using $\frac{P_{\text{test}}({\bf x})}{P_{\text{train}}({\bf x})}$.
Note that this is first proposed and proven as Theorem 3 in \citep{liu2014robust}. Here we derive using the generalized form to produce the same result. So when the assumption does not hold, the classification performance suffers. That motivates to use $P_{\text{gen}_v}$ instead as in (\ref{eq:robust}), which sits in between the most conservative and aggressive model and provides flexibility to find the right generalization distribution for training data. The good generalization distribution should help avoid errors brought by over optimistic estimates but also achieve better performance than no generalization.

\subsection{The Role of the Generalization Distribution} \label{sec:role}
Feature generalization makes it possible to utilize information shared by 
both training and testing distributions and is essential to improve performance
in predicting testing data. 
In order to illustrate the effect of the generalization distribution, in Figure \ref{fig:general} we consider a synthetic example with data sampled from two Gaussian distributions in 2-dimensional input space, with the training distribution totally contained in the testing distribution (two magenta ellipses). We compare the performance of logloss-based classifier on them. 100 data points are sampled and only the training datapoints are shown in Figure \ref{fig:general}, with 5\% of noise in both training and testing data. We assume there exists a generalization distribution that training features generalize to (white ellipse). After training with larger and larger generalization distribution, predictive performance is evaluated on testing data and logloss is shown under the figures. 

We can see from the figures that the generalization from training features get broader and broader with lager and larger generalization distribution. In the first case, $P_{\text{gen}_v}({\bf x})$ equals $P_{\text{train}}({\bf x})$, the method is equivalent with robust covariate shift method and the prediction is limited only to the space around where there is enough training support. In the second and third case, the certain portion in the whole space increases with logloss on testing data getting better. Finally, in the last case, $P_{\text{gen}_v}({\bf x})$ equals to $P_{\text{test}}({\bf x})$, the method is equivalent to importance weighting. We can see the prediction is quite certain across the whole space in this setting. The logloss, however, gets worse in this case due to the noise in the data. So the takeaway from this example is that it is important to get a balance between feature generalization and robustness.



{
\setlength\tabcolsep{0pt}
\begin{figure*}[htb] 
\begin{center}
\begin{tabular}{cccc}
\includegraphics[height=4.1cm, trim  = 4cm 0cm 7cm 0.8cm, clip]{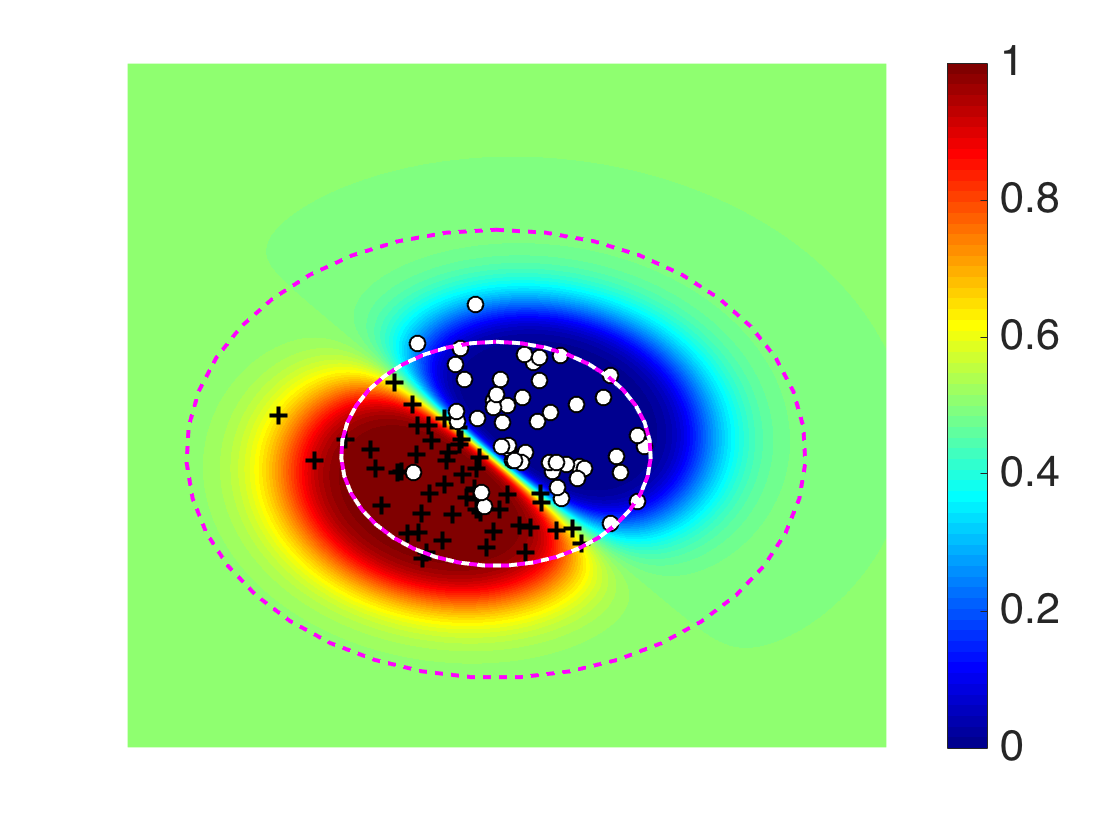}&
\includegraphics[height=4.1cm, trim  = 4cm 0cm 7cm 0.8cm, clip]{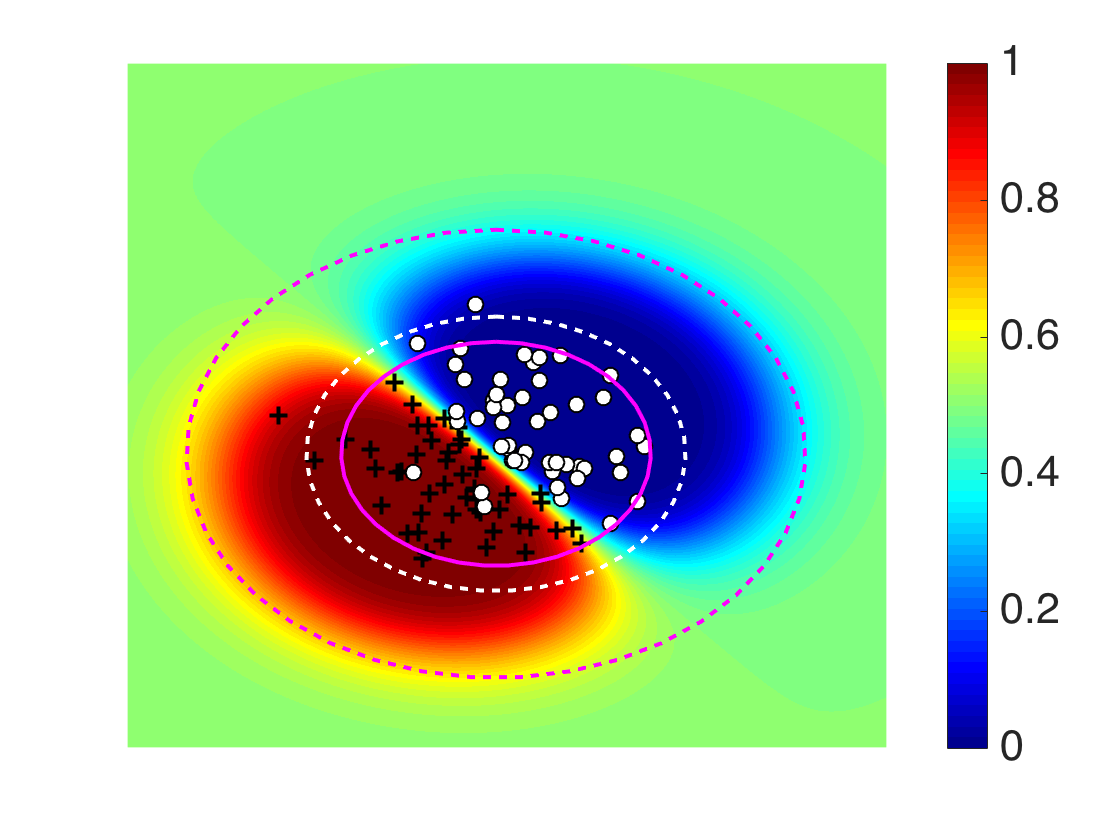} &
\includegraphics[height=4.1cm, trim  = 4cm 0cm 7cm 0.8cm, clip]{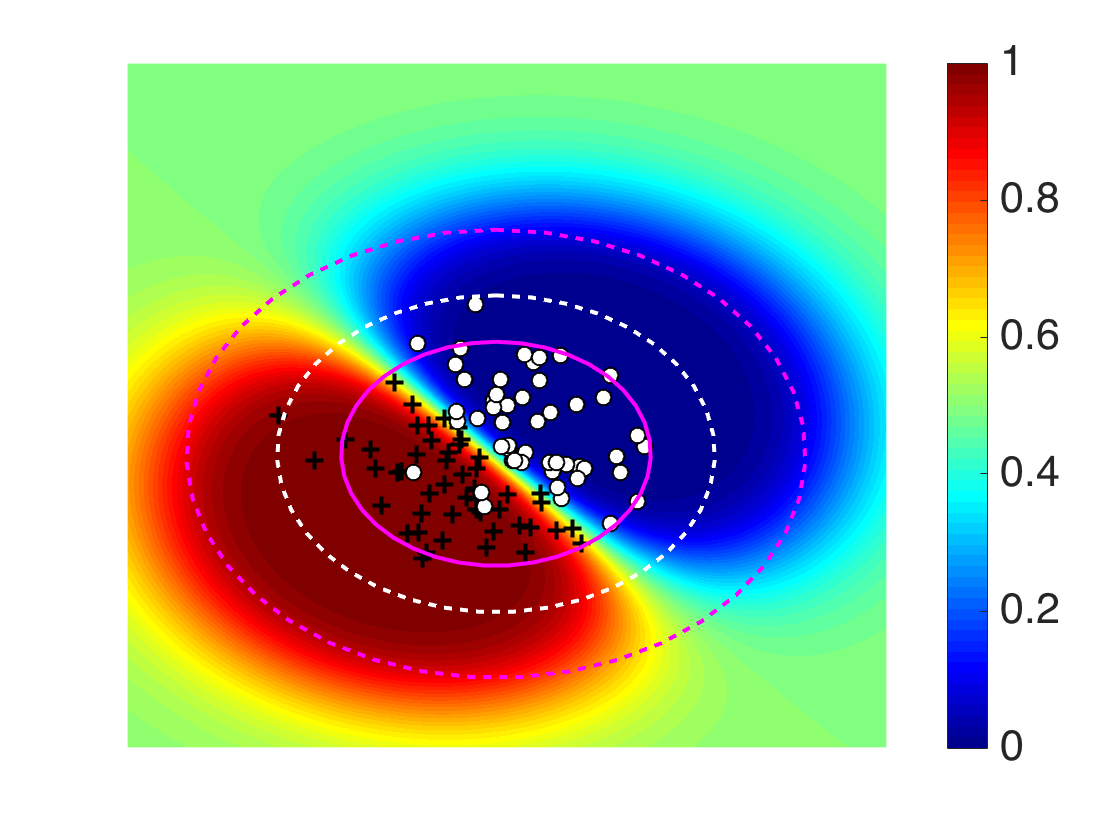} &
\includegraphics[height=4.1cm, trim  = 4cm 0cm 2cm 0.8cm, clip]{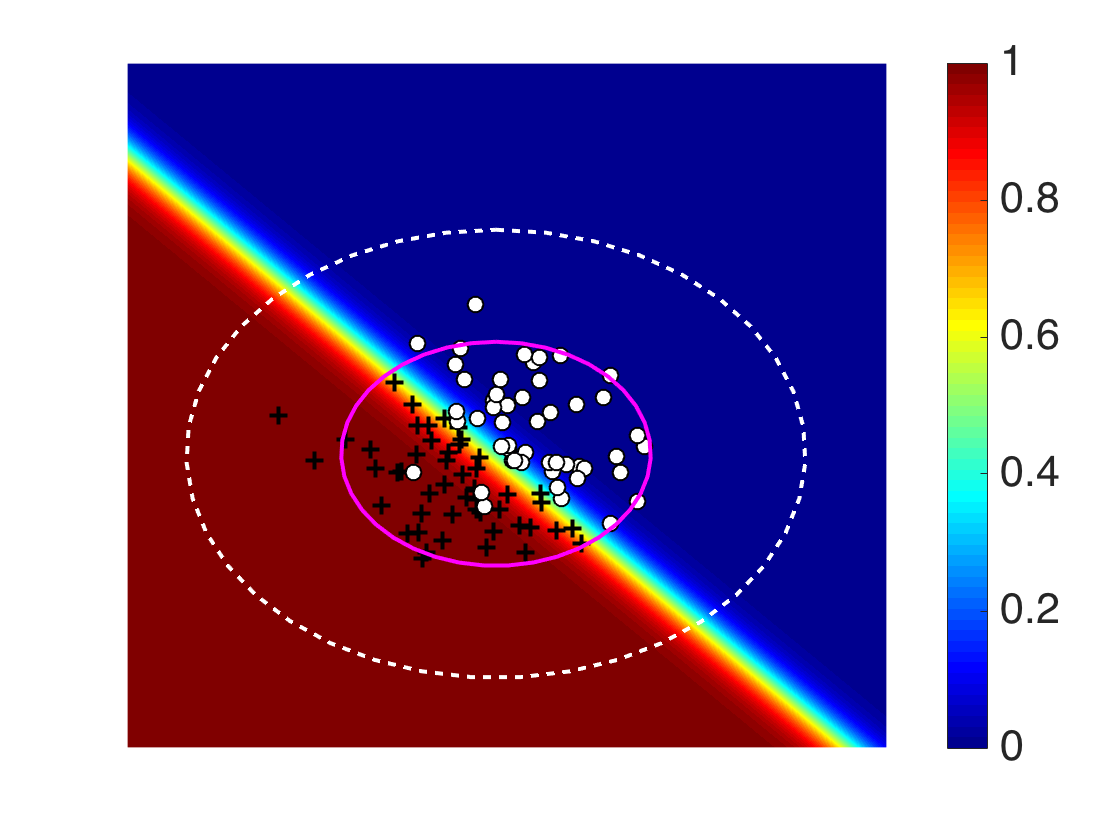} \\ 
logloss = 0.56 & logoss = 0.48 & logloss = 0.46& logloss = 0.75
\end{tabular}
\caption{
{\bf Comparison of incorporating different generalization distribution (white ellipses) in robust covariate shift classifier. Logloss evaluated on testing data points (not shown) is shown below each figure. Colormap represents the predicted probability of $P($`+ '$ |{\bf x})$. }
}
\label{fig:general}
\end{center}
\vspace{-4mm}
\end{figure*}
}

\subsection{View-based Feature Generalization} 
The choice of the generalization distribution contributes heavily to the resulting prediction on testing data. We now propose a possible set of assumptions for the generalized formulation in the case when there are multiple views of features.
We then explicitly apply assumptions about how each individual view of feature will 
generalize to the testing distribution. We denote the variables outside of view $v$ as ${\bf x}_{-v}$.
If we assume that certain view-based features partially generalize from the training 
distribution to the testing distribution by assuming only the input variables 
outside
of view $v$ generalize to testing distribution, this corresponds with the following relationships between inputs:
\begin{align}
& P_{\text{gen}_v}({\bf x}_{-v}|{\bf x}_v) = P_{\text{test}}({\bf x}_{-v}|{\bf x}_v) \notag\\
& P_{\text{gen}_v}({\bf x}_v) = P_{\text{train}}({\bf x}_v). \label{eq:assumption}
\end{align}

 Applying the above assumptions in the generalized formulation, the
right hand side of the constraints for those generalized views take the form of an importance weighting of view $v$'s feature
vector based on the non-view input variables, ${\bf x}_{-v}$: 
\begin{align}
\mathbb{E}_{({\bf X},Y) \sim \tilde{P}_{\text{train}}}
\left[\frac{P_{\text{test}}({\bf X}_{-v}|{\bf X}_v)}
{P_{\text{train}}({\bf X}_{-v}|{\bf X}_v)}
\phi_v({\bf X}_v,Y) \right]. \label{eq:features}
\end{align}

We use these partially reweighted features for those generalize features to formulate a new predictor
for classification under covariate shift in \S\ref{sec:formulation}.
This view-based robust classifier leverages partial generalization of features, which is possible in many applications when there exists noise or covariate shift in only certain feature views. This provides a solution that is robust to shift in a subset of feature views but also utilizes the information from the ones that are not shifted. 

\subsection{Robust Multiview Reformulation}
\label{sec:formulation}

Leveraging the view-based feature generalizations of Equation
\eqref{eq:features}, we re-formulate the adversarial game in Definition \ref{def:game} with set
$\mathcal{V}_g$ of generalized views and set $\mathcal{V}_o$ of 
non-generalized views of features.

\begin{definition} \label{def:game}
The {\bf robust multiview covariate shift classifier} is the solution to
the adversarial loss optimization game:
\begin{align}
& \min_{\hat{P}} \max_{\check{P}}
\mathbb{E}_{{\bf X} \sim P_{\text{test}}}
\left[\text{Loss}(\hat{P}_{\bf X}, \check{P}_{\bf X})\right].\label{eq:multiview}\\
& \text{ such that: } 
\forall v \in \mathcal{V}_g, \notag\\
& \quad \qquad\mathbb{E}_{{\bf X} \sim \tilde{P}_{\text{train}}, \check{Y}|{\bf X} \sim
\check{P} }
\left[\frac{P_{\text{test}}({\bf X}_{-v}|{\bf X}_v)}
{P_{\text{train}}({\bf X}_{-v}|{\bf X}_v)}
\phi_v({\bf X}_v,\check{Y}) \right] 
\notag\\
& \quad \qquad = 
\mathbb{E}_{({\bf X},Y) \sim \tilde{P}_{\text{train}}}
\left[\frac{P_{\text{test}}({\bf X}_{-v}|{\bf X}_v)}
{P_{\text{train}}({\bf X}_{-v}|{\bf X}_v)}
\phi_v({\bf X}_v,Y) \right], \notag\\
& \text{ and for: } \forall v' \in \mathcal{V}_o, \notag\\
&\qquad \qquad \qquad \qquad \mathbb{E}_{{\bf X} \sim \tilde{P}_{\text{train}}, \check{Y}|{\bf X} \sim
\check{P} }
\left[
\phi_{v'}({\bf X}_{v'},\check{Y}) \right] 
\notag\\
& \qquad \qquad \qquad \qquad 
= 
\mathbb{E}_{({\bf X},Y) \sim \tilde{P}_{\text{train}}}
\left[
\phi_{v'}({\bf X}_{v'},Y) \right]. \notag
\end{align}
\end{definition}
This definition implies that there are two different sets of constraints:
one set for features that we believe could be generalized,  and the one set
for features that we believe could not. Solving this constrained game formulation based on minimax duality and the method of Lagrangian multipliers for solving convex optimization problems, we have the parametric form of Theorem \ref{th:multiview} for conditional label probability distribution when $\text{Loss}(\check{P}_{\bf X}, \hat{P}_{\bf X}) = \mathbb{E}_{\check{Y}|{\bf X} \sim \check{P}} [-\log \hat{P}(\check{Y}|{\bf X})]$.
\begin{theorem} \label{th:multiview}The robust multiview covariate shift classifier when minimizing expected logloss has the
following parametric form:%
{\small
\begin{align}
&\hat{P}_\theta(y|{\bf x}) \propto \label{eq:multipara}\\
&\qquad
e^{\sum_v \frac{P_{\text{train}}({\bf x}_v)}{P_{\text{test}}({\bf x}_v)} 
\theta_v \cdot \phi_v({\bf x}_v, y) +  \frac{P_{\text{train}}({\bf x})}{P_{\text{test}}({\bf x})} \sum_{v'}
\theta_{v'} \cdot \phi_{v'}({\bf x}_{v'}, y)}
,\notag
\end{align}}%
where view-specific density ratios, $P_{\text{train}}({\bf x}_v)/P_{\text{test}}({\bf x}_v)$ are applied on the generalized views $\mathcal{V}_g$ and joint density ratios $P_{\text{train}}({\bf x})/P_{\text{test}}({\bf x})$
are applied on non-generalized views $\mathcal{V}_o$.
\end{theorem}
We show in the next theorem that in the logloss case, the parameter estimation for $\theta$ is equivalent with maximizing the conditional likelihood of testing data with $\hat{P}(Y|X)$ defined as \eqref{eq:multipara}. 
Therefore, the parameter can be estimated by using 
a gradient descent algorithm outlined in Theorem \ref{th:equivalent}
by using reweighted training samples.
\begin{theorem}\label{th:equivalent}
The parameters of the robust multiview covariate shift classifier are 
obtained through implicitly maximizing the conditional likelihood
of testing data by taking gradient steps as:
\begin{align}
&\mathbb{E}_{{\bf X} \sim \tilde{P}_{\text{train}}, \check{Y}|{\bf X} \sim
\check{P} }
\Bigg[\sum_v \frac{P_{\text{test}}({\bf X}_{-v}|{\bf X}_v)}
{P_{\text{train}}({\bf X}_{-v}|{\bf X}_v)}
\phi_v({\bf X}_v,\check{Y})  \notag\\
& \qquad
+ \sum_{v'} 
\phi_{v'}({\bf X}_{v'},\check{Y}) \Bigg] \notag\\
& -
\sum_v \mathbb{E}_{({\bf X},Y) \sim \tilde{P}_{\text{train}}}
\Bigg[\frac{P_{\text{test}}({\bf X}_{-v}|{\bf X}_v)}
{P_{\text{train}}({\bf X}_{-v}|{\bf X}_v)}
\phi_v({\bf X}_v,Y)\notag\\ &\qquad  + \sum_{v'} 
\phi_{v'}({\bf X}_{v'},\check{Y})\Bigg].\label{eq:gradient}
\end{align}
\end{theorem}

Note that even though we discuss the logloss case here in more detail, the same generalization assumption could also be applied to other loss functions. For those losses, there may not exist analytic forms for $\hat{P}$, but the subgradient should follow the same form as in (\ref{eq:gradient}). Therefore, as long as we are able to find an equilibrium of the inner minimax game in (\ref{eq:obj}), we can solve the optimization by subgradient descent.
\subsection{Understanding the Multiview Classifier}
We consider an illustrative synthetic example with data sampled from two overlapping Gaussian distributions ($X$) and identical true decision boundary ($Y$)
in Figure \ref{fig:result_logloss_1}. 
In 50 training and 100 testing data points, $10\%$ of the example are chosen
uniformly at random to be noise (label flipped). 
We train four methods, all of which are logloss-based,  using training data points (shown in the figures, roughly within the smaller ellipses) and evaluate them on testing data (not shown in the figures, roughly within larger ellipse).
The colormap represents the testing conditional label distribution in the whole space. Logloss evaluated on the testing data is listed below each figure.

{
\setlength\tabcolsep{0pt}
\begin{figure*}[tbh] 
\begin{center}
\begin{tabular}{cccc}
(a) LR& (b) IW & (c) Robust & (d) Robust - View\\
\includegraphics[height=4.1cm, trim  = 4cm 0cm 7cm 0.8cm, clip]{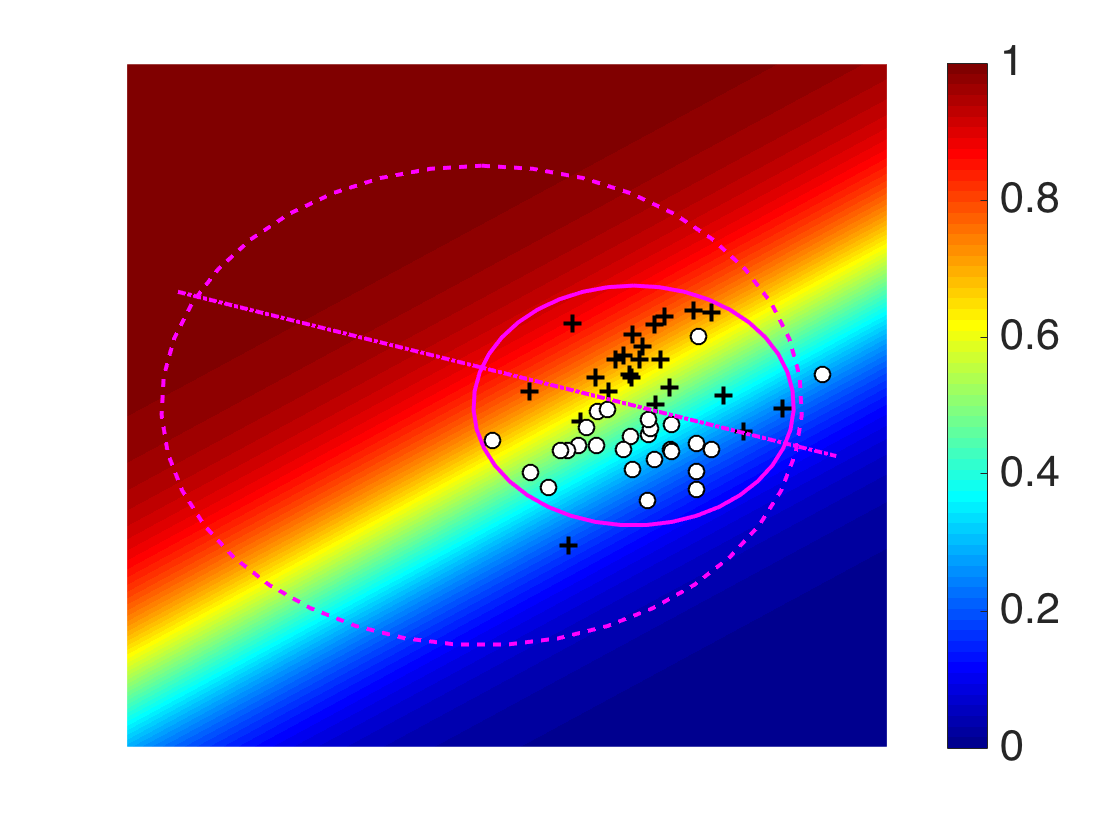}&
\includegraphics[height=4.1cm, trim  = 4cm 0cm 7cm 0.8cm, clip]{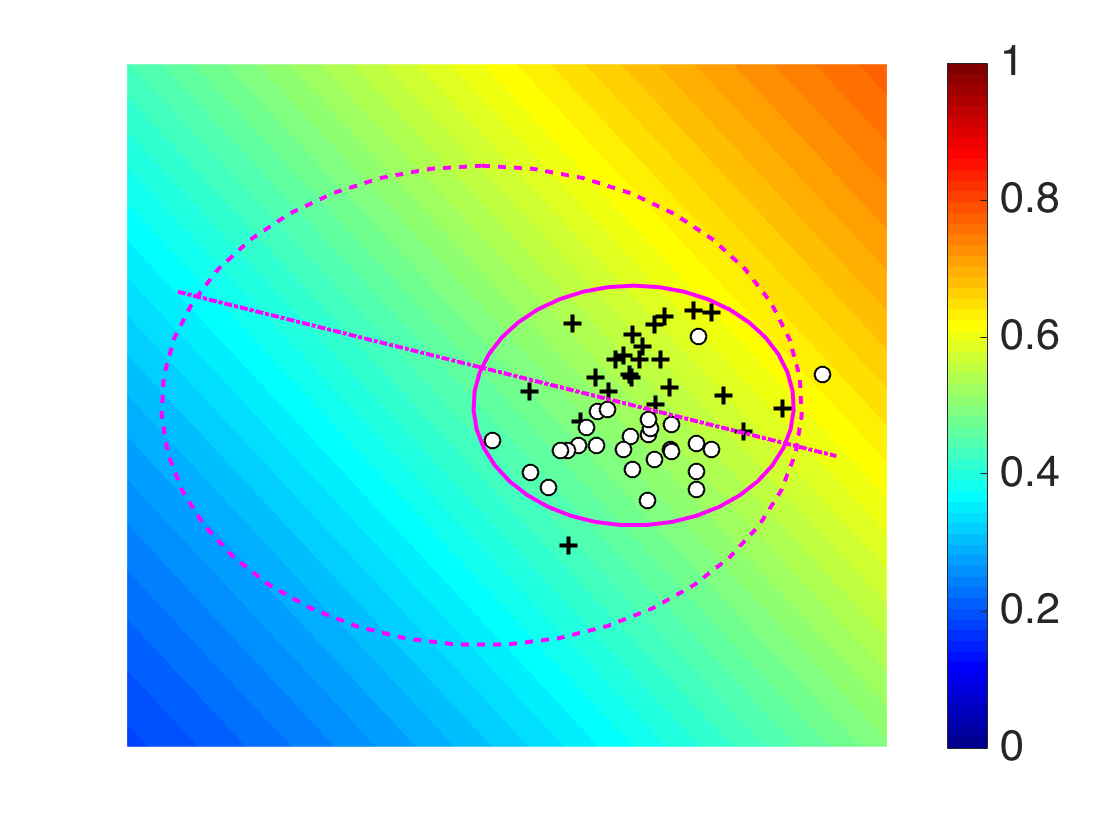} &
\includegraphics[height=4.1cm, trim  = 4cm 0cm 7cm 0.8cm, clip]{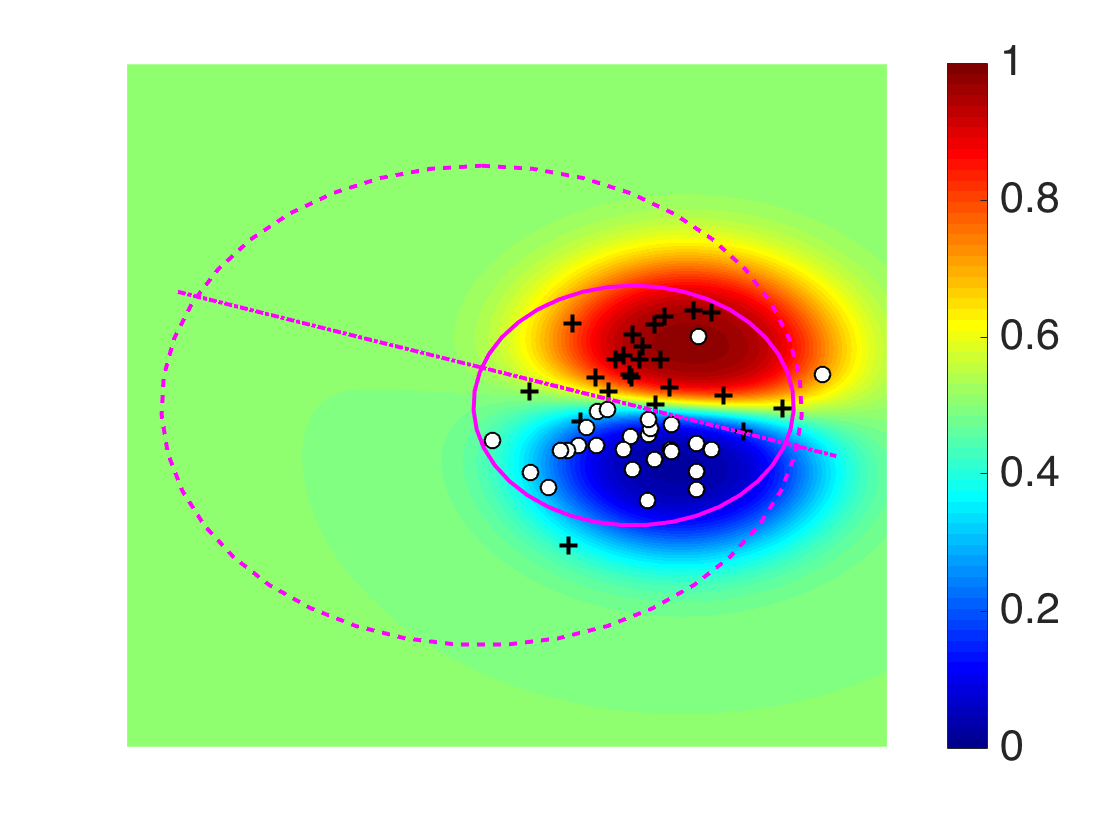} &
\includegraphics[height=4.1cm, trim  = 4cm 0cm 2cm 0.8cm, clip]{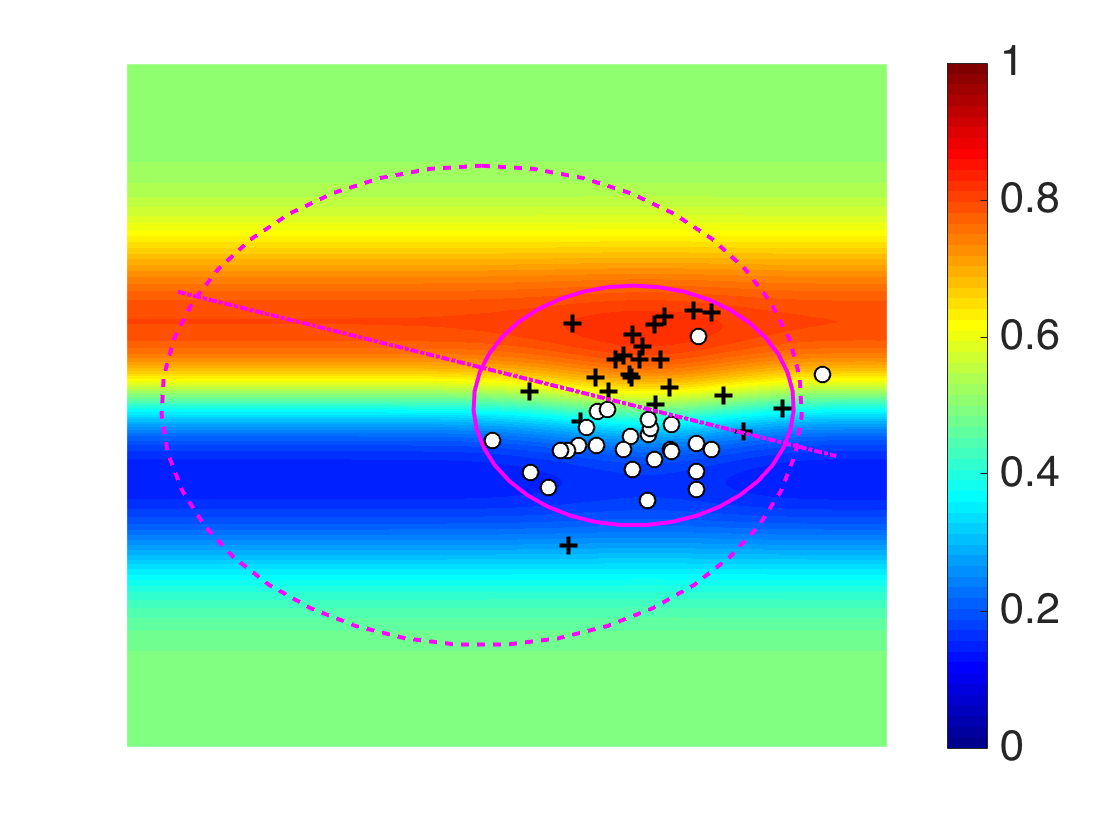} \\ 
logloss = 1.18 & logoss = 0.916 & logloss = 0.859 & logloss = 0.729
\end{tabular}
\caption{
{\bf Comparison of Logistic Regression (a), Importanct Weighting Logistic Regression (b), Robust Bias-Aware Prediction (c) and View-based Robust Bias-Aware Prediction (d). Logloss evaluated on testing data points (not shown) is shown below each figure. Colormap represents the predicted probability of $P($`+' $ |{\bf x})$. }
}
\label{fig:result_logloss_1}
\end{center}
\vspace{-4mm}
\end{figure*}
}

We see from the figures that the true decision boundary (the tilted line) could not be recovered by any of the methods using the limited data points. In fact, this is why covariate shift 
problems are so challenging, even though the assumption $P_{\text{train}}(y|x) = P_{\text{test}}(y|x)$ holds. LR makes very certain predictions based on the training data, but produces an incorrect decision boundary and a worse
logloss. IW, with reweighted training data, provides a less abrupt decision boundary but remains very certain towards the corners of the input space. The robust method, on the other hand, restricts the 
certain prediction regions only to areas with enough training data to support the prediction. The rest of the testing distribution space is covered with uniform predictions. It achieves better testing logloss by being more conservative. However, the question remains: could we leverage more information from the training data? We get our answer from the last model: our robust view-based model, which leverages
the fact that the view-based training and testing feature distribution is much closer in the vertical dimension (${\bf x}_2$) than in the horizontal dimension (${\bf x}_1$). Thus, the assumption that the training vertical feature dimension can generalize to the testing distribution in our generalized robust covariate shift classifier. This corresponds with a parametric form of $\hat{P}_\theta(y|{\bf x}) \propto
e^{\frac{P_{\text{train}}({\bf x})}{P_{\text{test}}({\bf x})} 
\theta_1 \cdot \phi_v({\bf x}_1, y) + \frac{P_{\text{train}}({\bf x}_2)}{P_{\text{test}}({\bf x}_2)} 
\theta_2 \cdot \phi_v({\bf x}_2, y)}$. 
The partial generalization robust method makes it possible to produce a solution that leverages the benefits of both the conservative robust method and the IW method. 
It maintains uncertainty in areas that have too little data to make predictions with any certainty (the top and bottom area in the input space), but gives more meaningful predictions in areas where the method expects the data could provide reasonable extrapolations.

\subsection{Bounding Expected Worst Case Test Loss} \label{sec:bound}
One significant difference between the robust covariate shift methods in this paper and empirical risk minimization based methods is that we directly minimize the worst case expected loss under the target distribution. The reason why this works is that the (sub-)gradient in our formulation only depends on the training distribution, so we are able to use training data to approximate it. On the contrary, the ERM based methods directly approximate the expected loss function using limited data as in Eq.\eqref{eq:reweight}. Despite this difference, we can easily control the error in our (sub-)gradients and therefore bound the error in the optimized worst case expected target loss.

We first define the notation $\text{WCLoss}(\theta)$ as the (regularized) worst case loss under the testing distribution, which is equivalent with the Lagrangian form of the optimization game of robust covariate shift classifier in (\ref{eq:obj}). Note that $\text{WCLoss}(\theta)$ differs in meaning from the $\text{Loss}(\hat{P}, \check{P})$ we used to optimize in the original framework in Definition (\ref{eq:generalgame}). For example, in logloss case, the worst case testing loss is obtained from the solved parametric form of $\hat{P}$ and $\check{P}$, which is the worst case predictor $P_{\theta}(\hat{Y}|{\bf X})$ as in (\ref{eq:robust}), to the $\text{Loss}(\hat{P}, \check{P})$: $\mathbb{E}_{P_{\text{test}(X, \hat{Y})}} [-\log P_{\theta}(\hat{Y}|{\bf X})]$. 
\begin{theorem}\label{th:bound}
Assuming we have $m$ training samples and $n$ dimensional features, the Lagrangian form of the robust covariate shift classifier (\ref{eq:obj}) is strong convex in terms of $\theta$ with strong convexity constant $M$, all density estimation is accurate, and the inner minimax game in (\ref{eq:obj}) is solved exactly, the expected loss on testing distribution of the robust covariate shift classifier 
is bounded, with probability $1 - \delta$:
\begin{align}
&\mathbb{E}_{P_{\text{test}}({\bf X})}[\text{WCLoss}(\hat{\theta})] \notag 
\le \mathbb{E}_{P_{\text{test}}({\bf X})}[\text{WCLoss}(\theta^*)]  + \frac{n \log\frac{2n}{\delta}}{4Mm}.
\end{align}
\end{theorem}
This bound indicates the distance between the expected target loss induced by our learned model from $m$ training data and the optimal target loss is decreasing with speed $\mathcal{O}(\frac{1}{m})$. Note that the strong convexity condition is easy to satisfy even with non-smooth loss functions with $L_2$ regularization.  We refer to proof in the supplemental materials.

\section{Experiments}
We conduct experiments on real datasets and investigate the performance of two classifiers from the robust covariate shift general framework. One corresponds with using 0-1 loss in the framework but just considers all features as one view and assumes $P_{\text{gen}_v}({\bf x}) = P_{\text{train}}({\bf x})$. The other uses logloss in (\ref{eq:multiview}) and applies generalization assumptions (\ref{eq:assumption}) on multiview features. We chose four datasets from the UCI repository \citep{UCI, dataset_vehicle} for both sets of experiments. We show the detailed information about each dataset in supplement \ref{sup:datasets}.  In order to create covariate shift, we
 synthetically  
generate 30 
separate 
experiments in each dataset by drawing 
100 training samples and 100 testing data samples 
from it, following similar sampling procedure in \cite{huang2006correcting}, which is described in supplement \ref{sup:sampling}. Note that we normalize the data to $[0, 1]$ beforehand. For each method in both sets of experiments, the regularization weights are chosen by 5-fold cross validation or importance weighted cross validation(IWCV). We refer to the details in supplement \ref{sup:model}.

\subsection{Logistic Regression as Density Estimator}
We use a discriminative classifier---logistic regression---as the density (ratio)
estimator.
To estimate ratios like $\frac{P_{\text{train}}({\bf X})}{P_{\text{test}}({\bf X})}$, according to the Bayes' Rule:
$\frac{P_{\text{train}}({\bf X})}{P_{\text{test}}({\bf X})} = \frac{P({\bf X}|\text{train})}{P({\bf X}|\text{test})} = \frac{P(\text{train}|X)}{P(\text{test}|X)}\frac{P(\text{test})}{P(\text{train})}$.
To estimate density ratios as 
$\frac{P_{\text{test}}({\bf X}_{-v}|{\bf X}_v)}
{P_{\text{train}}({\bf X}_{-v}|{\bf X}_v)}$, we analyze this ratio according to Bayes' rule and obtain:
$\frac{P_{\text{test}}({\bf X}_{-v}|{\bf X}_v)}
{P_{\text{train}}({\bf X}_{-v}|{\bf X}_v)}
= \frac{P({\text{test}}|{\bf X})}{P({\text{train}}|{\bf X})} \cdot
\frac{P({\text{train}}|{\bf X}_v )}{P( {\text{test}}|{\bf X}_v)}$.
Therefore, it is easy to construct two logistic regression models to estimate each ratio for ${\bf X}$ and ${\bf X}_v$ respectively. We carefully control the magnitude of the density ratio by adjusting the regularization
using finite sample properties of the training data. 
According to \citep{dudik2006maximum}, the amount of regularization corresponds
to the slack of feature expectation matching in constraints in the primal form of maximum entropy type of problems, like logistic regression. 
If we are using $L_2$ regularization and 
$\lambda = D_2 [1+(2+\sqrt{2})\sqrt{\ln(1/\sigma)} )] / \sqrt{2m}$,
where $D_2$ is the $l_2$ diameter of the feature space, the loss of the estimated conditional label distribution will be bounded as 
$L(\hat{\theta}) \le
L(\theta^*) + ||\theta^*||_2 D_2 [\sqrt{2} + 2(1+\sqrt{2})\sqrt{\ln(1/\sigma)}]/\sqrt{m}$,
with probability at least $1-\sigma$. 
Therefore, we use the maximum range among all the features to estimate $D_2$ in the regularization for density estimation in both robust and importance weighted methods. Note that since we normalize features, $D_2 \le 1$ and $\sigma$ is set to 0.05. We use first order features for density estimation in our experiment, which is enough in most cases.

\subsection{Robust 0-l Loss Classifier}
We evaluate four methods:
\begin{itemize}[topsep=0pt,itemsep=-1ex,partopsep=0ex,parsep=1ex]
\item {\bf Robust bias aware 0-1 classifier (Robust 0-1)} utilizes the general robust covariate shift classification framework (\ref{eq:generalgame}) with $\text{Loss}(\check{P}_{\bf X}, \hat{P}_{\bf X}) = \hat{P}^{T}C\check{P}$ with $C$ as the 0-1 loss matrix and $P_{\text{gen}_v}({\bf x}) = P_{\text{train}}({\bf x})$.
\item {\bf Adversarial 0-1 classifier (Adv 0-1)} minimizes expected 0-1 loss on the training distribution and has an optimization objective of: $\min_{\theta}\min_{\hat{P}} \max_{\check{P}}
\mathbb{E}_{{\bf X} \sim P_{\text{train}}}
[\hat{P}^{T}C\check{P} + \sum_v\theta_v\phi_v(X_v, \check{Y})]$ $- \sum_v\theta_v\tilde{\phi_v} + \epsilon ||\theta||_2$, where $\tilde{\phi} =   \mathbb{E}_{({\bf X},Y) \sim \tilde{P}_{\text{train}}}
\left[
\phi_v({\bf X}_v,Y) \right]$ here.
\item {\bf Multiclass SVM (SVM)} follows \cite{crammer2001algorithmic} by minimizing hinge loss on training data.
\item {\bf Importance Weighted SVM (IW-SVM)} reweights the training data with $\frac{P_{\text{test}}(X)}{P_{\text{train}}(X)}$ and minimizes reweighted hinge loss on training data.
\end{itemize}
We show the comparison of accuracy in Table \ref{tab:0_1_accuracy} and highlight methods that are either the best under paired t-test or not statistically distinguishable with significance level 0.1 in bold. We can see that Robust 0-1 performs better than other methods except in \verb|Seed|, where it is statistically no worse than others. And Robust 0-1 can improve from Adv 0-1 at most times. That means minimizing worst case test loss using the adversarial game formulation (\ref{eq:generalgame}) under covariate shfit is better than minimizing training loss and ignoring the bias using the same formulation.

{
\setlength\tabcolsep{3pt}
\begin{table}[htb]
\begin{center}
\caption{Average Accuracy Comparison for UCI datasets} \label{tab:0_1_accuracy}
{\small
\begin{tabular}{|K{1.6cm}|K{1.5cm} |K{1.3cm}|K{1.3cm}|K{1.3cm}|}
\hline 
{\bf Datasets}&{\bf Robust 0-1}  &{\bf Adv 0-1}  & {\bf SVM} & {\bf IW-SVM}\\
\hline
\verb|Seed|  &{\bf 0.834} & {\bf 0.820}&{\bf0.820} & {\bf 0.820}\\
\hline
\verb|Vertebral|& {\bf 0.823} & 0.805 & 0.748 & 0.748\\
\hline 
\verb|Vehicle| &{\bf 0.547} &  0.535 & 0.497& 0.497      \\
\hline
\verb|Spam| & {\bf 0.757} & 0.711 &0.724 & 0.724 \\
\hline
\end{tabular}
}
\end{center}
\vspace{-5mm}
\end{table}
}
\subsection{Robust Multiview Covariate Shift Classifier}
For a second set of experiments, we regard each feature dimension as a specific view to simplify our experimental setup for UCI datasets. We use KL-divergence as the criterion to determine the features that are generalizable or not. Details for the criterion is in supplement \ref{sup:criterion}. Besides the UCI datasets, we also evaluate our method on the multi-view dataset \verb|Language| \citep{amini2009learning}, which consist of text features of documents in five different languages (English-EN, French-FR, Germany-GR, Italian-IT, and Spanish-SP). This dataset is generated by translating documents originally in one language to the other four languages using machine translation. We regard different language features as different views for this task. In our experiment, we use the document originally in English. We use two languages in training and testing, with one view the same and the other view different between training and testing. 

There are six categories as labels, more than ten thousand features for each language and around twenty thousand samples for English documents. To better estimate the densities we use PCA to reduce the dimension of features to 100 for each view and randomly sample 500 datapoints as training and 500 datapoints as testing.
Therefore, we construct different settings for this dataset. For example, we can train using English and French views and test on Germany and French views (\verb|EN FR - GR FR|). 
We evaluate the multiview robust covariate shift approach and three other methods: 
%
\begin{itemize}[topsep=0pt,itemsep=-1ex,partopsep=0ex,parsep=1ex]
\item {\bf Multiview robust bias aware classifier (Robust - View)} utilize the general robust covariate shift classification framework applying multiview feature generalization assumptions as in Definition \ref{def:game}.
\item {\bf Robust bias aware classifier (Robust)} adversarially
minimizes the testing distribution logloss Eq.\ref{eq:value}, using the parametric form as Eq.\ref{eq:robust}.
\item {\bf Logistic regression (LR)} 
maximizes the conditional log likelihood on training data, 
$\max_{\theta} \mathbb{E}_{P_{\text{train}}(x)P{(y|x)}} \left[\log P_{\theta} (Y|X) \right] - \lambda \| \theta \|_2$,
where $ \hat{P}_{\theta} (y|x) = \frac{\exp(\theta \cdot \Phi(x,y))}{\sum_{y' \in \mathcal{Y}} \exp(\theta \cdot \Phi(x,y'))} $ and $\lambda$ is the regularization constant.   
This approach ignores the covariate shift of the problem setting entirely. 
\item {\bf Importance weighting method (IW)} 
maximizes the conditional testing data likelihood as estimated using importance
weighting 
with the density ratio, 
$\max_{\theta} \mathbb{E}_{P_{\text{train}}(x)P(y|x)} \left[  \frac{P_{\text{test}}(x)}{P_{\text{train}}(x)} \left(\log P_{\theta} (Y|X)  \right)\right] - \lambda \| \theta \|_2$.
\end{itemize}
{
\setlength\tabcolsep{3pt}
\begin{table}[htb]
\begin{center}
\caption{Average Logloss Comparison for UCI datasets} \label{tab:result_average_1}
{\small
\begin{tabular}{|K{2cm}|K{1.8cm}|K{1cm}|K{1cm}|K{1cm}|}
\hline 
{\bf Dataset}&{\bf Robust-View}  &{\bf Robust}  & {\bf LR} & {\bf IW} \\
\hline
\verb|Seed| &  {\bf 1.039} & {\bf 1.105} &1.385 & 1.299  \\
\hline
\verb|Vertebral| & {\bf 0.577} & 0.830 & 0.811 & 0.810 \\
\hline 
\verb|Vehicle| &  {\bf 1.68} & 1.82 & 2.82 & 2.59 \\
\hline
\verb|Spam| & {\bf 0.853} & 1.804 &1.981 & 0.969 \\
\hline
\end{tabular}
}
\end{center}
\vspace{-5mm}
\end{table}
}

{
\setlength\tabcolsep{3pt}
\begin{table}[htb]
\begin{center}
\caption{Logloss Comparison for Language datasets} \label{tab:result_average_2}
{\small
\begin{tabular}{|K{2.5cm}|K{1.8cm}|K{1cm}|K{1cm}|K{1cm}|}
\hline 
{\bf Dataset}&{\bf Robust-View}  &{\bf Robust}  & {\bf LR} & {\bf IW} \\
\hline
\verb|EN FR - GR FR| &1.88&2.44&11.04&10.39
\\
\hline
\verb|EN GR - FR GR|&1.69&2.38&6.53&6.15
\\
\hline 
\verb|IT GR - FR GR| & 1.96&2.48&8.40&7.59  \\
\hline
\verb|EN IT - GR IT|  & 1.94 &2.54 & 12.72 & 8.31  \\
\hline
\end{tabular}
}
\end{center}
\vspace{-5mm}
\end{table}
}
We compare logloss of each method in Table \ref{tab:result_average_1} and Table \ref{tab:result_average_2}. We denote the significantly best result under paired t-test with significance level 0.05 in bold numbers for UCI datasets. 
We can see from the Table \ref{tab:result_average_1} that Robust - View outperforms all other methods in most datasets for UCI experiments by having the lowest logloss. Moreover, it always improve from Robust, except being comparable with Robust in \verb|Seed| for logloss and in \verb|Vehicle| for accuracy. On the other hand, the performance of the other methods are mixed, with Robust achieves slightly better logloss and comparable accuracy with LR and IW.  

For \verb|Language| datasets, LR and IW are actually even worse than the Random baseline in terms of logloss due to the possibly large shift between different languages. In contrast, Robust -View and Robust are usually better than the baseline due to their robustness property. And Robust -View can improve from Robust in both logloss by utilizing the generalization property of certain features, especially when Robust is even worse than IR in accuracy because it is overly uncertain with logloss close to random. The reason why Robust is so close to uniform is that it regards all features as a whole and differentiate training and testing data. It disregards the fact that there are useful information that could be used to improve the predictive performance, which is exactly what motivates this work.

\section{Conclusion}
Covariate shift classification 
is an important but difficult task for machine learning in non-stationary environments when the testing labels are not available. The original robust bias-aware classifier only optimize logloss and may lose predictive power on testing input space that lacks training support, especially when dimension of features are large. We propose a general robust covariate shift classification framework that is flexible enough to minimize various loss functions and make different feature generalization assumptions for multiview features. We derive different classifiers from the framework and demonstrate a set of assumptions that could be applied to achieve a balance of robustness and informativeness. We use synthetic examples, UCI biased datasets and real multiview datasets to demonstrate the model performance. We also investigate the theoretical property of the framework and are able to bound the worse case testing loss of our model properly.

\bibliography{biblio}
\bibliographystyle{plainnat}
\newpage
\input{supplement}

\end{document}

%% file: supplement.tex
\appendix
\onecolumn

\section{Supplement}

\subsection{Proof for Theorem \ref{th:general}}
\label{pf:general}
\begin{proof}
Solving the constrained minimax game \eqref{eq:generalgame}, we follow \cite{grunwald2004game} and \cite{liu2014robust}. The minimax game reduces to a constrained maximum entropy problem:
\begin{align}
& \max_{\hat{P}}
\mathbb{E}_{{\bf X} \sim P_{\text{test}},
\hat{Y}|{\bf X} \sim \hat{P}}
\left[-\log \hat{P}(\hat{Y}|{\bf X})\right].\\
& \text{ such that: } \forall v \in \{1, \hdots, |\mathcal{V}|\}, 
\mathbb{E}_{{\bf X} \sim \tilde{P}_{\text{gen}}, \hat{Y}|{\bf X} \sim
\hat{P} }
\left[
\phi_v({\bf X}_v,\hat{Y}) \right] 
= 
\mathbb{E}_{({\bf X},Y) \sim \tilde{P}_{\text{train}}}
\left[\frac{P_{\text{gen}}(X)}{P_{\text{train}}(X)}
\phi_v({\bf X}_v,Y) \right], \notag\\ &
\qquad\forall x \in \mathcal{X}, \mathbb{E}_{\hat{Y}|{\bf X} \sim
\hat{P} } [1|X] = 1, \qquad 
\forall x \in \mathcal{X}, y\in \mathcal{Y}: \hat{P}(y|x) \ge 0.
\end{align}
Solving this constrained optimization problem using Lagrangian multiplier method, the Lagrangian is:
\begin{align}
&\mathcal{L}(\hat{P}(y|x), \theta, \lambda({\bf x})) = \mathbb{E}_{{\bf X} \sim P_{\text{test}},
\hat{Y}|{\bf X} \sim \hat{P}}
\left[-\log \hat{P}(\hat{Y}|{\bf X})\right] 
+ \theta \cdot (\mathbb{E}_{{\bf X} \sim \tilde{P}_{\text{gen}}, \hat{Y}|{\bf X} \sim
\hat{P} }
\left[
\phi_v({\bf X}_v,\hat{Y}) \right] \notag\\
&\qquad - \mathbb{E}_{({\bf X},Y) \sim \tilde{P}_{\text{train}}}
\left[\frac{P_{\text{gen}}(X)}{P_{\text{train}}(X)}
\phi_v({\bf X}_v,Y) \right]) +  
\sum_{x\in \mathcal{X}} \lambda({\bf x})[\mathbb{E}_{\hat{Y}|{\bf X}  \sim \hat{P}}[1|X] - 1].
\end{align}
According to strong Lagrangian duality, the max and min could be switched and preserving the same value:
\begin{align}
\max_{\hat{P}(y|{\bf x})} \min_{\theta, \lambda({\bf x})} \mathcal{L}(\hat{P}(y|{\bf x}), \theta,\lambda({\bf x})) = \min_{\theta, \lambda}\max_{\hat{P}(y|{\bf x})}\mathcal{L}(\hat{P}(y|{\bf x}), \theta, \lambda({\bf x})).
\end{align}
So, assuming a fixed $\theta$ and $\lambda({\bf x})$, the internal maximization problem can be solved first. Taking the partial derivative with respect to the conditional probability of a specific $y$ and $x$, $\hat{P}(y|{\bf x})$,
\begin{align}
&\frac{\partial}{\partial \hat{P}(y|{\bf x})}\mathcal{L}(\hat{P}(y|{\bf x}), \theta,\lambda({\bf x})) = -P_{\text{test}}({\bf x})\log\hat{P}(y|{\bf x})
- P_{\text{test}}({\bf x}) + P_{\text{gen}}({\bf x})\sum_v\theta\cdot \phi_v({\bf x}_v,y) + \lambda({\bf x}),
\end{align}
setting it equal to zero, and solving it, we obtain:
\begin{align}
\log\hat{P}(y|x) = -1 + \frac{P_{\text{gen}}({\bf x})}{P_{\text{test}}({\bf x})}\sum_v \theta\cdot\phi_v({\bf x}_v,y) + \frac{\lambda({\bf x})}{P_{\text{test}}({\bf x})}.
\end{align}
Therefore,
\begin{align}
\hat{P}(y|x) =e^{ -1 + \frac{P_{\text{gen}}({\bf x})}{P_{\text{test}}({\bf x})}\sum_v \theta \cdot \phi_v({\bf x}_v,y) + \frac{\lambda({\bf x})}{P_{\text{test}}({\bf x})}}.
\end{align}
We analytically solve the normalization terms and yielding the conditional probability distribution:
\begin{align}
\hat{P}(y|{\bf x}) = \frac{e^{\frac{P_{\text{gen}}({\bf x})}{P_{\text{test}}({\bf x})}\sum_v\theta\cdot \phi_v({\bf x}_v,y)}}{Z_{\theta}({\bf x})},
\end{align}
where $Z_{\theta}({\bf x}) = \sum_{y'\in \mathcal{Y}}e^{\frac{P_{\text{gen}}({\bf x})}{P_{\text{test}}({\bf x})}\sum_v\theta \cdot \phi_v({\bf x}_v,y')}$.
\end{proof}

\subsection{Proof for Theorem  \ref{th:multiview}}
\label{pf:multiview}
\begin{proof}
Solving the constrained minimax game \eqref{def:game}, we follow a similar procedure as in Proof \ref{pf:general}. The minimax game reduces to a constrained maximum entropy problem:

\begin{align}
& \min_{\hat{P}} \max_{\hat{P}}
\mathbb{E}_{{\bf X} \sim P_{\text{test}},
\hat{Y}|{\bf X} \sim \hat{P}}
\left[-\log \hat{P}(\hat{Y}|{\bf X})\right].\\
& \text{ such that: } 
\forall v \in \mathcal{V}_g, 
\mathbb{E}_{{\bf X} \sim \tilde{P}_{\text{train}}, \hat{Y}|{\bf X} \sim
\hat{P} }
\left[\frac{P_{\text{test}}({\bf X}_{-v}|{\bf X}_v)}
{P_{\text{train}}({\bf X}_{-v}|{\bf X}_v)}
\phi_v({\bf X}_v,\hat{Y}) \right] 
= 
\mathbb{E}_{({\bf X},Y) \sim \tilde{P}_{\text{train}}}
\left[\frac{P_{\text{test}}({\bf X}_{-v}|{\bf X}_v)}
{P_{\text{train}}({\bf X}_{-v}|{\bf X}_v)}
\phi_v({\bf X}_v,Y) \right], \notag\\
& \text{ and for: } \forall v' \in \mathcal{V}_o, 
\mathbb{E}_{{\bf X} \sim \tilde{P}_{\text{train}}, \hat{Y}|{\bf X} \sim
\hat{P} }
\left[
\phi_{v'}({\bf X}_{v'},\hat{Y}) \right] 
= 
\mathbb{E}_{({\bf X},Y) \sim \tilde{P}_{\text{train}}}
\left[
\phi_{v'}({\bf X}_{v'},Y) \right] \notag\\
&\qquad\qquad \forall  x \in \mathcal{X} \mathbb{E}_{\hat{Y}|{\bf X} \sim
\hat{P} } [1|X] = 1\qquad
\forall x \in \mathcal{X}, y\in \mathcal{Y}: \hat{P}(y|x) \ge 0.
\end{align}
Solving this constrained optimization problem using Lagrangian multiplier method, the Lagrangian is:
\begin{align}
&\mathcal{L}(\hat{P}(y|x), \theta, \lambda({\bf x})) = \mathbb{E}_{{\bf X} \sim P_{\text{test}},
\hat{Y}|{\bf X} \sim \hat{P}}
\left[-\log \hat{P}(\hat{Y}|{\bf X})\right] 
+ \theta_v \cdot (\mathbb{E}_{{\bf X} \sim \tilde{P}_{\text{train}}, \hat{Y}|{\bf X} \sim
\hat{P} }
\left[
   \frac{P_{\text{test}}(X_v|X_{-v})}{P_{\text{train}}(X_v|X_{-v})}\phi_v({\bf X}_v,\hat{Y}) \right]  \notag\\
&\qquad- \mathbb{E}_{({\bf X},Y) \sim \tilde{P}_{\text{train}}}
\left[\frac{P_{\text{test}}(X_v|X_{-v})}{P_{\text{train}}(X_v|X_{-v})}
\phi_v({\bf X}_v,Y) \right]) +  
\theta_{v'} \cdot (\mathbb{E}_{{\bf X} \sim \tilde{P}_{\text{train}}, \hat{Y}|{\bf X} \sim
\hat{P} }
\left[
  \phi_{v'}({\bf X}_{v'},\hat{Y}) \right]  \notag\\
&\qquad - \mathbb{E}_{({\bf X},Y) \sim \tilde{P}_{\text{train}}}
\left[
\phi_{v'}({\bf X}_{v'},Y) \right]) 
+ \sum_{x\in \mathcal{X}} \lambda({\bf x})[\mathbb{E}_{\hat{Y}|{\bf X}  \sim \hat{P}}[1|X] - 1].
\end{align}
Taking the partial derivative with respect to the conditional probability of a specific $y$ and $x$, $\hat{P}(y|{\bf x})$,
\begin{align}
&\frac{\partial}{\partial \hat{P}(y|{\bf x})}\mathcal{L}(\hat{P}(y|{\bf x}), \theta,\lambda({\bf x})) = -P_{\text{test}}({\bf x})\log\hat{P}(y|{\bf x})
- P_{\text{test}}({\bf x}) + \sum_v P_{\text{train}}({\bf x}_v) P_{\text{test}}({\bf x}_{-v}|x_v)\theta_v\cdot \phi_v({\bf x}_v,y)\notag\\
&\qquad\qquad +\sum_{v'} P_{\text{train}}({\bf x})\theta_{v'}\cdot \phi_{v'}({\bf x}_{v'},y)+ \lambda({\bf x}),
\end{align}
setting it equal to zero, and solving it, we obtain:
\begin{align}
\log\hat{P}(y|x) = -1 +\sum_v \frac{P_{\text{train}}({\bf x}_v)}{P_{\text{test}}({\bf x}_v)}\theta_v\cdot\phi_v({\bf x}_v,y)  + \sum_{v'}\frac{P_{\text{train}}({\bf x})}{P_{\text{test}}({\bf x})}\theta_{v'}\cdot \phi_{v'}({\bf x}_{v'}, y) + \frac{\lambda({\bf x})}{P_{\text{test}}({\bf x})}.
\end{align}
Therefore,
\begin{align}
\hat{P}(y|{\bf x}) =e^{ -1 + \sum_v \frac{P_{\text{train}}({\bf x}_v)}{P_{\text{test}}({\bf x}_v)}\theta_v\cdot\phi_v({\bf x}_v,y) +\sum_{v'}\frac{P_{\text{train}}({\bf x})}{P_{\text{test}}({\bf x})}\theta_{v'} \cdot \phi_{v'}({\bf x}_{v'},y) + \frac{\lambda({\bf x})}{P_{\text{test}}({\bf x})}}.
\end{align}
We analytically solve the normalization terms and yielding the conditional probability distribution:
\begin{align}
\hat{P}(y|{\bf x}) = \frac{e^{\sum_v\frac{P_{\text{train}}({\bf x}_v)}{P_{\text{test}}({\bf x}_v)}\theta_v\cdot\phi_v({\bf x}_v,y)+ \sum_{v'}\frac{P_{\text{train}}({\bf x})}{P_{\text{test}}({\bf x})}\theta_{v'}\cdot \phi_{v'}({\bf x}_{v'},y)}}{Z_{\theta}({\bf x})},
\end{align}
where $Z_{\theta}({\bf x}) = \sum_{y'\in \mathcal{Y}}e^{\sum_v \frac{P_{\text{train}}({\bf x}_v)}{P_{\text{test}}({\bf x}_v)}\theta_v\cdot\phi_v({\bf x}_v,y')+ \sum_{v'}\frac{P_{\text{train}}({\bf x})}{P_{\text{test}}({\bf x})}\theta_{v'} \cdot \phi_{v'}({\bf x}_{v'},y')}$.
\end{proof}

\subsection{Proof for Theorem \ref{th:equivalent}}
\begin{proof}
Plugging in the parametric form of $\hat{P}(y|x)$ into
the Lagrangian objective function in Proof \ref{pf:multiview}, we have:
\begin{align}
& \theta^*= \argmax_\theta
\mathbb{E}_{({\bf X}, Y) \sim P_{\text{train}}}[ 
\sum_v \frac{P_{\text{test}}({\bf X}_{-v}|{\bf X}_v)}
{P_{\text{train}}({\bf X}_{-v}|{\bf X}_v)} 
\theta_v \cdot \phi_v({\bf X}_v, Y) \\
&\qquad+ \sum_{v'}
\theta_{v'} \cdot \phi_{v'}({\bf X}_{v'}, Y)]\notag\\
& \qquad-\mathbb{E}_{{\bf X} \sim P_{\text{test}}}\Bigg[
\log  \sum_{y' \in \mathcal{Y}}
e^{\sum_v \frac{P_{\text{train}}({\bf X}_v)}{P_{\text{test}}({\bf X}_v)} 
\theta_v \cdot \phi_v({\bf X}_v, y') + \sum_{v'} \frac{P_{\text{train}}({\bf X})}{P_{\text{test}}({\bf X})} 
\theta_{v'}  \cdot \phi_{v'}({\bf X}_{v'}, y')}
 \Bigg] \\
& = \argmax_\theta
\mathbb{E}_{({\bf X}, Y) \sim P_{\text{test}}}\Bigg[
\sum_v \frac{P_{\text{train}}({\bf X}_v)}{P_{\text{test}}({\bf X}_v)} 
\theta_v \cdot \phi_v({\bf X}_v, Y) \notag\\
& \qquad + \sum_{v'} \frac{P_{\text{train}}({\bf X})}{P_{\text{test}} ({\bf X})}\theta_{v'}\phi_{v'}({\bf X}_{v'}, Y) \notag\\
&\qquad -\log  \sum_{y' \in \mathcal{Y}}
e^{\sum_v \frac{P_{\text{train}}({\bf X}_v)}{P_{\text{test}}({\bf X}_v)} 
\theta_v \cdot \phi_v({\bf X}_v, y') + \sum_{v'} \frac{P_{\text{train}}({\bf X})}{P_{\text{test}}({\bf X})} 
\theta_{v'}  \cdot \phi_{v'}({\bf X}_{v'}, y')}
 \Bigg] \notag \\
&= \argmax_\theta
\mathbb{E}_{({\bf X}, Y) \sim P_{\text{test}}}\left[\log 
\hat{P}_\theta(Y|{\bf X}) \right] 
\end{align}
\end{proof}

\subsection{Proof for Theorem \ref{th:bound}}
\label{sup:bound}
\begin{proof}
We first investigate the empirical approximation of (sub-)gradient $\tilde{G}$ and see how far it could deviate from the true (sub-)gradient $G$. 
\begin{align}
| ||G||^2 - ||\tilde{G}||^2 | 
& = | ||\mathbb{E}_{({\bf X}, Y)\sim P_{train}}[\phi(X, Y)] - \tilde{\phi}||^2 - ||\mathbb{E}_{({\bf X}, Y)\sim \tilde{P}_{train}}[\phi(x, y)] - \tilde{\phi}||^2 |,\notag\\
&\le || \mathbb{E}_{({\bf X}, Y)\sim P_{train}}[\phi(x, y)]  - \mathbb{E}_{({\bf X}, Y)\sim \tilde{P}_{train}}[\phi(x, y)] ||^2\notag\\
&=\sum_{i = 1}^n |\mathbb{E}_{({\bf X}, Y)\sim P_{train}}[\phi_i(x, y)]- \mathbb{E}_{({\bf X}, Y)\sim \tilde{P}_{train}}[\phi_i(x, y)]|^2, 
\end{align}
where  $n$ is the total dimension of features in $\phi(x,y)$. 
According to Hoeffding bound:
\begin{align}
P(|\mathbb{E}_{({\bf X}, Y)\sim P_{train}}[\phi_i(x, y)]- \mathbb{E}_{({\bf X}, Y)\sim \tilde{P}_{train}}[\phi_i(x, y)] |> \epsilon) < 2e^{-2n\epsilon^2}
\end{align} 
Then we have the following, with probability $1-\delta$, 
\begin{align}
&| ||G||^2 - ||\tilde{G}||^2 |  \le \frac{nlog\frac{2n}{\delta}}{2m}.
\end{align}
The reason we are interested in the error in norm-2 of (sub-)gradient is we want to utilize the property that for a strongly convex objective function the following is true:
\begin{align}
f(t) - min_{s\in S}f(s) \le \frac{1}{2M}||\triangledown f(t) ||^2,
\end{align}
where $M$ is the constant for strong convexity, i.e. $f(s) \ge f(t) + \triangledown f(t) ^T (s - t) + \frac{M}{2}||s-t||^2$. 
This is also true when the objective function is not smooth, when $ \triangledown f(t)$ can be replaced by subgradient $g \in \partial f(x)$. Therefore, if we assume $min_{s\in S}f(s)$, which in our case is the true worse case expected target loss, is reached at $||G||^2 = 0$, then the objective function is bounded by
\begin{align}
\mathbb{E}_{({\bf X}, Y)\sim P_{\text{test}}}[\text{Loss}(\hat{\theta})] \le \mathbb{E}_{({\bf X}, Y)\sim P_{\text{test}}}[\text{Loss}(\theta^*)]  + \frac{nlog\frac{2n}{\delta}}{4Mm},
\end{align}
with probability $1 - \delta$, where $\text{Loss}(\theta)$ is the worst case loss function in the general game formulation.
\end{proof}
\subsection{Detailed Information of Datasets}
\label{sup:datasets}{
\setlength\tabcolsep{3pt}
\begin{table}[h]
\begin{center}
\caption{Datasets information and sampling details for synthetically biased experiments} \label{tab:datasets_info_1}
{\small
\hspace*{-0.2cm}\begin{tabular}{|K{2cm}|K{1.5cm}|K{1.5cm}|K{1.5cm}|}
\hline 
{\bf Dataset} & {\bf Features}  & {\bf Examples} 
& {\bf Classes} \\
\hline
\verb|Seed| & 7 & 210 & 3 \\
\hline
\verb|Vertebral| & 6 & 310 & 3 \\
\hline
\verb|Vehicle| & 18 & 946 & 4 \\
\hline
\verb|Spam| & 57 & 4601& 2 \\

\hline
\end{tabular}
}
\end{center}
\end{table}
}

\subsection{Biased sampling strategy in UCI datasets}\label{sup:sampling}
\begin{enumerate}
\item Separate the data into training and testing portion according to a feature;
\item Perform Principal Component Analysis (PCA) on both portions of data respectively;
\item Generate a random value $a$ and $b$
from different intervals;
\item Randomly choose a principal component $i$, 
calculate the weight vector as $normpdf(m_i, \mu_i, \sigma_i)$, where $\mu_i = min(m_i) + (max(m_i) - min(m_i)) / a$, $\sigma_i = std(m_i) / b$;
\item Sample examples for the testing data samples from the testing portion of data in 
proportion to the weight vector values;
\item Follow the same procedure to sample examples for the training data.
\end{enumerate}

\subsection{Logistic regression as Density Estimation}\label{sup:density}
\begin{align}
&\frac{P_{\text{test}}({\bf X}_{-v}|{\bf X}_v)}
{P_{\text{train}}({\bf X}_{-v}|{\bf X}_v)} = \frac{P_{\text{test}}({\bf X})}{P_{\text{train}}({\bf X})} \cdot 
\frac{P_{\text{train}}({\bf X}_v)}{P_{\text{test}}({\bf X}_v)} \notag \\
&=  \frac{P({\bf X}|{\text{test}})}{P({\bf X}|{\text{train}})} \cdot 
\frac{P({\bf X}_v |{\text{train}})}{P({\bf X}_v |{\text{test}})} \notag\\
&=   \frac{P({\text{test}}|{\bf X}) P({\bf X})/ P(\text{test})}{P({\text{train}}|{\bf X}) P({\bf X})/ P(\text{train})} \notag\\
&\qquad \cdot
\frac{P({\text{train}}|{\bf X}_v ) P({\bf X}_v)/ P(\text{train})}{P( {\text{test}}|{\bf X}_v) P({\bf X}_v)/ P(\text{test})}\notag\\
& = \frac{P({\text{test}}|{\bf X})}{P({\text{train}}|{\bf X})} \cdot
\frac{P({\text{train}}|{\bf X}_v )}{P( {\text{test}}|{\bf X}_v)}.
\end{align}

\subsection{Generalization Criterion}
\label{sup:criterion}
For UCI datasets, we regard each feature dimension as a view. We evaluate the KL-divergence of the training distribution $P_{\text{train}}({\bf x}_v)$ and the testing distribution $P_{\text{test}}({\bf x}_v)$ after density estimation to determine whether we should assume the generalization of each view, i.e., $v \in \mathcal{V}_o$ or $ v \in \mathcal{V}_g$. We use the threshold of 0.1, that if $K < 0.1$, we consider $P_{\text{train}}({\bf x}_v)$ to be similar enough with $P_{\text{test}}({\bf x}_v)$ and $v \in \mathcal{V}_g$, otherwise,  $v \in \mathcal{V}_o$. We include both training and testing inputs in the computation of KL-divergence. 
\begin{align}
&K(P_{\text{train}}({\bf x}_v), P_{\text{test}}({\bf x}_v)) \notag\\
&= \sum_{{\bf x}_v \in {\bf x}_\text{train}} P_{\text{train}}({\bf x}_v) log (P_{\text{train}}({\bf x}_v)/P_{\text{test}}({\bf x}_v)) \notag\\
&+ \sum_{{\bf x}_v \in {\bf x}_\text{test}} P_{\text{test}}({\bf x}_v) log (P_{\text{test}}({\bf x}_v)/P_{\text{train}}({\bf x}_v))\notag
\end{align}

For \verb|Language| datasets, we assume we do not have prior knowledge for which view of features should be generalized. We conduct density estimation on both views and detect the one which is similar between training and testing. In practice, we could rely on both data observation and expert knowledge to choose the generalization criterion. 

\subsection{Model Selection}
\label{sup:model}
For each method, 
the regularization parameter $\lambda$ are chosen using 5-fold cross validation, or importance weighted cross validation (IWCV) on a parameter range $\lambda \in [2^{-16}, 2^{-12}, 2^{-8}, 2^{-4}, 1]$.
Here the traditional cross validation is applied on LR, while IWCV is applied on all the other methods.
Note that the traditional cross validation process is not correct anymore in the covariate shift setting where the training marginal data distribution of $P({\bf x})$ is
different from the testing distribution \citep{sugiyama2007covariate}. 
Therefore, standard cross validation only matches the logistic regression method which ignores the bias. Though IWCV was originally designed for the importance weighting methods, 
it is proven to be unbiased for any loss function.  We apply it to perform model tuning for our robust methods, even though the error estimate variance could 
be large.